\documentclass{article}

\usepackage{times}
\usepackage{graphicx} 
\usepackage{subfigure}

\usepackage{rotating}
\usepackage{array}
\usepackage{multirow}

\usepackage{natbib}

\usepackage{algorithm}
\usepackage{algorithmic}

\usepackage{hyperref}
\usepackage{breakurl}

\usepackage[accepted]{icml2014}

\icmltitlerunning{No more meta-parameter tuning in unsupervised sparse feature learning}

\begin{document}

\twocolumn[
\icmltitle{No more meta-parameter tuning\\ in unsupervised sparse feature learning}

\icmlauthor{Adriana Romero}{adriana.romero@ub.edu}
\icmladdress{Departament de Matem\`{a}tica Aplicada i An\`{a}lisi,
            Universitat de Barcelona, Barcelona, Spain.}
\icmlauthor{Petia Radeva}{petia.ivanova@ub.edu}
\icmladdress{Departament de Matem\`{a}tica Aplicada i An\`{a}lisi,
            Universitat de Barcelona, Barcelona, Spain.}
\icmlauthor{Carlo Gatta}{cgatta@cvc.uab.es}
\icmladdress{Centre de Visi\'{o} per Computador,
            Bellaterra, Spain.}

\icmlkeywords{unsupervised feature learning, pre-training of deep networks, sparse visual features}

\vskip 0.3in
]

\begin{abstract}
We propose a meta-parameter free, off-the-shelf, simple and fast unsupervised feature learning algorithm, which exploits a new way of optimizing for sparsity. Experiments on STL-10 show that the method presents state-of-the-art performance and provides discriminative features that generalize well.

\end{abstract}

\section{Introduction}
\label{sec:intro}
Significant effort has been devoted to handcraft appropriate feature representations of data in several fields. In tasks such as image classification and object recognition, unsupervised learned features have shown to compete well or even outperform manually designed ones \cite{Ranzato06,Yang09,Coates11_AISTATS}. Unsupervised feature learning has also shown to be helpful in greedy layerwise pre-training of deep architectures \cite{Hinton06,Bengio06,Larochelle09,Erhan10}.

In \cite{Bengio09}, the author claims that potentially interesting research involves pre-training algorithms, which ``[...] would be proficient at extracting good features but involving an easier optimization problem.'' In addition to that, one of the main criticisms to state-of-the-art methods is that they require a significant amount of meta-parameters \cite{Bengio13}. As stated in \cite{Snoek12}, the tuning of these meta-parameters is a laborious task that requires expert knowledge, rules of thumb or extensive search and, whose setting can vary for different tasks. Therefore, there is great interest for meta-parameter free methods \cite{Ngiam11} and automatic approaches to optimize the performance of learning algorithms \cite{Snoek12}.

Nevertheless, little effort has been devoted to address this problem (see Table \ref{tab:metaparams} for a comparison of meta-parameters required by unsupervised feature learning methods). To the best of our knowledge, work in this direction includes ICA \cite{Hyvarinen00,Hyvarinen01} and sparse filtering \cite{Ngiam11}. Although ICA provides good results at object recognition tasks \cite{Le11,Ngiam11}, the method scales poorly to large datasets and high input dimensionality.

Computational complexity is also a major drawback of many state-of-the-art methods. ICA requires an expensive orthogonalization to be computed at each iteration. Sparse coding has an expensive inference, which requires a prohibitive iterative optimization. Significant amount of work has been done in order to overcome this limitation \cite{Lee06,Kavukcuoglu10}. Predictive Sparse Decomposition (PSD) \cite{Kavukcuoglu10} is a successful variant of sparse coding, which uses a predictor to approximate the sparse representation and solves the sparse coding computationally expensive encoding step.

\begin{table*}
\caption{Meta-parameters to tune of state-of-the-art unsupervised feature learning methods.}
\small
\centering
\begin{tabular}{| c | c |}
\hline
\textbf{Method} & \textbf{Meta-parameters to tune}\\ \hline\hline
Sparse RBM \cite{Hinton06,Lee08} & \begin{tabular}{@{}c@{}}weight decay, sparseness constant, \\ sparsity penalty, momentum\end{tabular}\\ \hline
\begin{tabular}{@{}c@{}}Sparse \\ auto-encoders \cite{Ranzato06}\end{tabular}  & \begin{tabular}{@{}c@{}}weight decay, sparseness constant, \\ sparsity penalty\end{tabular}\\ \hline
Sparse Coding \cite{Olshausen97} & sparsity penalty\\ \hline
RICA \cite{Le11} & reconstruction penalty\\ \hline
PSD \cite{Kavukcuoglu10} & sparsity penalty, prediction penalty\\ \hline
OMP-k \cite{Pati93,Blumensath07,Coates11_ICML} & $k$ (non-zero elements)\\ \hline
ICA \cite{Hyvarinen00,Hyvarinen01} & - \\ \hline
Sparse Filtering \cite{Ngiam11} & - \\ \hline
\end{tabular}
\label{tab:metaparams}
\end{table*}

In this paper, we aim to solve some of the above-mentioned problems. We propose a \textbf{meta-parameter free, off-the-shelf, simple and fast} approach, which exploits a new way of optimizing for a sparsity, without explicitly modeling the data distribution. The method iteratively builds an \textit{ideally} sparse target and optimizes the dictionary by minimizing the error between the system output and the \textit{ideally} sparse target. \textit{Defining sparsity concepts in terms of expected output allows to exploit a new strategy in unsupervised training. }

It is worth stressing that many optimization strategies can be used to minimize the above-mentioned error and that parameters of these optimization techniques must not be considered as belonging to our approach.

Experiments on STL-10 dataset show that the method outperforms state-of-the-art methods in single layer image classification, providing discriminative features that generalize well.

Linear feature extraction methods combined with sparse coding encodings are among best performers on object recognition datasets. The importance of properly combining training/encoding and encoding/pooling strategies has been argued in \cite{Coates11_ICML} and \cite{Zeiler13} respectively. Since the goal of this paper is to propose a new method for unsupervised feature learning, dealing with all the possible combinations of encoding and pooling could mask the benefits of the method that we propose. However, for the sake of fair comparison with the state-of-the-art, we test the method with sparse coding and soft-threshold encodings combined with sum pooling, following the experimental pipeline of \cite{Coates11_ICML}.

\section{State-of-the-art}
\label{sec:soa}
Commonly used algorithms for unsupervised feature learning include Restricted Boltzmann Machines (RBM) \cite{Hinton06}, auto-encoders \cite{Bengio06}, sparse coding \cite{Raina07} and hybrids such as PSD \cite{Kavukcuoglu10}. Many other methods such as ICA \cite{Hyvarinen00,Hyvarinen01}, Reconstruction ICA (RICA) \cite{Le11}, Sparse Filtering \cite{Ngiam11} and methods related to vector quantization such as Orthogonal Matching Pursuit (OMP-k) \cite{Pati93,Blumensath07,Coates11_ICML} have also been used in the literature to extract unsupervised feature representations. These algorithms could be divided into two categories: explicitly modeling or not the input distribution. Sparse auto-encoders \cite{Ranzato06}, sparse RBM \cite{Hinton06,Lee08,Hinton10,Goh12}, sparse coding \cite{Olshausen97}, PSD \cite{Kavukcuoglu10}, OMP-k \cite{Pati93,Blumensath07,Coates11_ICML} and Reconstruction ICA (RICA) \cite{Le11} explicitly model the data distribution by minimizing the reconstruction error. Although learning a good approximation of the data distribution may be desirable, approaches such as sparse filtering \cite{Ngiam11} show that this seems not so important if the goal is to have a discriminative sparse system. Sparse filtering does not attempt to explicitly model the input distribution but focuses on the properties of the output distribution instead.

Sparsity is among the desirable properties of a good output representation \cite{Field94,Olshausen97,Ranzato06,Lee08,Le11,Ngiam11,Bengio13}. Sparse features consist of a large amount of outputs, which respond rarely and provide high responses when they do respond. Sparsity can be described in terms of population sparsity and lifetime sparsity \cite{Willmore01}. Both lifetime and population sparsity are important properties of the output distribution. On one hand, lifetime sparsity plays an important role in preventing bad solutions such as numerous dead outputs. There seems to be a consensus to overcome such degenerate solutions, which is to ensure similar statistics among outputs \cite{Field94,Willmore01,Ranzato06,Ngiam11}. On the other hand, population sparsity helps providing a simple interpretation of the input data such as the ones found in early visual areas. To the best of our knowledge, the definition of population sparsity remains ambiguous.

State-of-the-art methods optimize either for one or both sparsity forms in their objective function. The great majority seeks sparsity using the $L_1$ penalty and does not optimize for an explicit level of sparsity in their outputs. Sparse auto-encoders optimize for a target activation allowing to deal with lifetime sparsity; nevertheless, the target activation requires tuning and does not explicitly control the level of population sparsity. OMP-k defines the level of population sparsity by setting $k$ to the maximum expected number of non-zero elements per output code, whereas the methods in \cite{Olshausen97,Ranzato06,Lee08,Le11,Ngiam11} do not explicitly define the proportion of outputs expected to be active at the same time.

\section{Method}
\label{sec:method}
In this section, we describe how the proposed method learns a sparse feature representation of the data in terms of population and lifetime sparsity. The method iteratively builds an \textit{ideally} sparse target and optimizes the dictionary by minimizing the error between the system output and the \textit{ideally} sparse target. Subsection \ref{ssec:SLS_SPS} highlights the algorithm to enforce lifetime and population sparsity in the ideally sparse target. Subsection \ref{ssec:opt} provides implementation details on the system and optimization strategies used to minimize the error between the system output and the ideally sparse target.

\subsection{Enforcing Population and Lifetime Sparsity by defining an ideal target}
\label{ssec:SLS_SPS}
We define population and lifetime sparsity as properties of an \textit{ideal} sparse output. Given $N$ training samples and an output of dimensionality $N_{h}$, we define the first property of the output as:
\begin{enumerate}
\item \textbf{Strong Lifetime Sparsity:} The output vectors must be composed solely of active and inactive units (no intermediate values between two fixed scalars are allowed) and all outputs must activate for an equal number of inputs. Activation is exactly distributed among the $N_{h}$ outputs.
\end{enumerate}
Our Strong Lifetime Sparsity definition is a more strict requirement than the high dispersal concept introduced in \cite{Ngiam11}, since they only require that ``the mean squared activations of each feature (output) [...] should be roughly the same for all features (outputs)''. While high dispersal attempts to diversify the learned bases, it does not guarantee the output distribution, in the lifetime sense, to be composed of only a few activations. Furthermore, our definition ensures the absence of dead outputs.

Given our definition of Strong Lifetime Sparsity, the population sparsity must require that, for each training sample, only one output element is active:
\begin{enumerate}
\setcounter{enumi}{1}
\item \textbf{Strong Population Sparsity:} For each training sample only one output must be active.
\end{enumerate}

The rationale of our approach is to appropriately generate an ideal output target that fulfils properties (1) and (2), and then learn the parameters of the system by minimizing the $L_{2}$ error between the output target and the output generated by the system during training. In this way, \textit{we seek a system optimized for both population and lifetime sparsity in an explicit way.}

The key component of our approach is how to define the ideal output target based on the above-mentioned properties. However, to ensure that the optimization of the system parameters converges, we add a third property:

\begin{enumerate}
\setcounter{enumi}{2}
\item \textbf{Minimal Perturbation:} The ideal output target should be defined as the best approximation of the system output by means of $L_2$ error fulfilling properties (1) \& (2).
\end{enumerate}

Creating the output target that ensures the above-mentioned properties is analogous to solving an assignment problem. The Hungarian method \cite{Kuhn55} is a combinatorial optimization algorithm, which solves the assignment problem. However, its computational cost $\mathcal{O}((N N_h)^{3/2})$ is prohibitive. Therefore, in the next section we propose a simple and fast $\mathcal{O}(N N_{h})$ algorithm to generate the ideal output target, which ensures sparsity properties (1) and (2) and provides an approximate solution for minimal perturbation property (3).

\subsubsection{Ideal target generation: the Enforcing Population and Lifetime Sparsity (EPLS) algorithm}
\label{sssec:EPLS}

Let us assume that we have a system, which produces a row output vector $\mathbf{h}$. We use the notation $\mathbf{h}_{j}$ to refer to one element of $\mathbf{h}$. We define an output matrix $\mathbf{H}$ composed of $N_b$ output vectors of dimensionality $N_h$, such that $N_b \leq N$. Likewise, we define an ideal target output matrix $\mathbf{T}$ of the same size. Algorithm \ref{algorithm:idealTarget} details the EPLS algorithm to generate the \textit{ideal} target $\mathbf{T}$ from $\mathbf{H}$. For the sake of simplicity, every step of the algorithm where the subscript $j$ appears must be applied $\forall j \in \{1,2,\ldots,N_{h}\}$.

\begin{algorithm}
\caption{EPLS}
\label{algorithm:idealTarget}
\begin{algorithmic}[1]
\small
\REQUIRE $\mathbf{H}$, $\mathbf{a}$, N
\ENSURE $\mathbf{T}$, $\mathbf{a}$
\STATE $\mathbf{T} = 0$
\FOR{$n = 1 \to N_b$}
\STATE $\mathbf{h}_{j} = \mathbf{H}_{n,j}$
\STATE $k = \arg \max_{j} \left(\mathbf{h}_{j} - \mathbf{a}_{j}\right)$
\STATE $\mathbf{T}_{n,k} = 1$
\STATE $\mathbf{a}_{k} = \mathbf{a}_{k} + \frac{N_h}{N} + \epsilon$
\ENDFOR
\STATE Remap $\mathbf{T}$ to active/inactive values of the corresponding function.
\end{algorithmic}
\end{algorithm}
Starting with no activation in $\mathbf{T}$ (line 1), the algorithm proceeds as follows. A row vector $\mathbf{h}$ from $\mathbf{H}$ is processed at each iteration (line 3). The crucial step is performed in line 4: the output $k$ that has to be activated in the $n^{th}$ row of $\mathbf{T}$ is selected as the one that has the maximal activation value $\mathbf{h}_j$ minus the inhibitor $\mathbf{a}_j$. The inhibitor $\mathbf{a}_j$ can be seen as an accumulator that ``counts" the number of times an output $j$ has been selected, increasing its inhibition progressively by $N_h/N$ until reaching maximal inhibition. This prevents the selection of an output that has already been activated $N/N_{h}$ times. The rationale behind the equation in line 4 is that, while selecting the maximal responses in the matrix $\mathbf{H}$, we have to take care to distribute them evenly among all outputs (in order to ensure Strong Lifetime Sparsity). Using this strategy, it can be demonstrated that the resulting matrix $\mathbf{T}$ perfectly fulfills properties (1) and (2). In line 5, the algorithm activates the $k^{th}$ element of $n^{th}$ row of the target matrix $\mathbf{T}$. By activating the ``relative" maximum, we approximate property (3). Finally, the inhibitor $\mathbf{a}$ is updated in line 6.


\subsection{System and Optimization strategies}
\label{ssec:opt}
Let us assume that we have a system parameterized by $\Gamma = \{\mathbf{W}, \mathbf{b}\}$, with activation function $f$, which takes as input a data vector $\mathbf{d}$ and produces an output vector $\mathbf{h} = f(\mathbf{d},\Gamma)$. We use the same notation as in Section \ref{sec:method} and define a data matrix $\mathbf{D}$ composed of $N$ rows and $N_{d}$ columns, where $N_{d}$ is the input dimensionality.

To compare our training strategy to previous well known systems, we tested our algorithm using
\begin{equation}
\mathbf{H} = f \left(\mathbf{D} \mathbf{W} + \mathbf{b} \right),
\label{eq:activation}
\end{equation}
where $f$ is a logistic non-linearity.

\subsubsection{Optimization strategy}
\label{sssec:stardardopt}

The system might be trained by means of an off-the-shelf mini-batch Stochastic Gradient Descent (SGD) method with adaptive learning rates such as variance-based SGD (vSGD) \cite{Schaul13}. Algorithm \ref{alg:standardEPLS} details the latter training process. The mini-batch size $N_b$ can be set to any value, in all the experiments we have set $N_b = N_h$. Starting with $\Gamma$ set to small random numbers as in \cite{LeCun98} (line 1), at each epoch we shuffle the samples of the training set (line 3), reset the EPLS inhibitor $\mathbf{a}$ to a flat activation (line 4) and process all mini-batches. For each mini-batch $b$, samples $\mathbf{D}^{(b)}$ are selected (line 6). Then, the output $\mathbf{H}^{(b)}$ is computed (line 7) and the EPLS is invoked to compute $\mathbf{T}^{(b)}$ and update $\mathbf{a}$ (line 8). After that, the gradient of the error is computed (line 9) and the learning rate $\eta$ is estimated as in \cite{Schaul13} (line 10). The system parameters are then updated to minimize the $L_{2}$ error $E^{(b)} = || \mathbf{H}^{(b)} - \mathbf{T}^{(b)} ||_{2}^2$ (line 11). Finally, the bases $\mathbf{W}$ in $\Gamma$ are limited to have unit norm to avoid degenerate solutions (line 13). This procedure is repeated until a stop condition is met; in our experiments, the training stops when the relative decrement error between epochs is small ($<10^{-6}$).

When updating the system parameters, we assume that $\mathbf{T}$ does not depend on $\Gamma$, thus $\frac{\partial \mathbf{T}}{\partial \Gamma} = 0$; we carried out experiments that show that this approximation does not significantly influence the gradient descent convergence nor the quality of the minimization. Moreover, this assumption makes the algorithm faster, since we remove the need of computing the numerical partial derivatives of $\mathbf{T}$.

The mini-batch vSGD allows to scale the algorithm easily, especially with respect to the number of samples $N$.

\begin{algorithm}
\caption{Standard EPLS training}
\label{alg:standardEPLS}
\begin{algorithmic}[1]
\small
\REQUIRE $\mathbf{D}$
\ENSURE $\Gamma$
\STATE $\Gamma =$ small random values
\REPEAT
\STATE Shuffle $\mathbf{D}$ randomly
\STATE $\mathbf{a} =$ flat activation
\FOR{$b = 1 \to \lfloor N/N_b \rfloor$}
\STATE Select mini-batch samples $\mathbf{D}^{(b)}$
\STATE $\mathbf{H}^{(b)} = f(\mathbf{D}^{(b)},\Gamma)$
\STATE $(\mathbf{T}^{(b)},\mathbf{a}) = EPLS(\mathbf{H}^{(b)},\mathbf{a},N)$
\STATE $\mathbf{G} = \nabla_{\Gamma}|| \mathbf{H}^{(b)} - \mathbf{T}^{(b)} ||_2^2$
\STATE Estimate learning rate $\eta$ as in \cite{Schaul13}
\STATE $\Gamma = \Gamma - \eta \mathbf{G}$
\ENDFOR
\STATE Limit the bases $\mathbf{W}$ in $\Gamma$ to have unit norm
\UNTIL stop condition verified
\end{algorithmic}
\end{algorithm}
%

\section{Experiments}
\label{sec:exp}

The performance of training and encoding strategies in single layer networks has been extensively analyzed in the literature \cite{Coates11_AISTATS,Coates11_ICML,Ngiam11} on STL-10\footnote{http://www.stanford.edu/$\sim$acoates/stl10/} dataset. STL-10 dataset consists of 96x96 pixels color images belonging to 10 different classes. The dataset is divided into a large unlabeled training set containing 100K images and smaller labeled training and test sets, containing 5000 and 8000 images, respectively. It has to be considered that in STL-10, the primary challenge is to make use of the unlabeled data (100K images), which is 100 times bigger than the labeled data used to train the classifier (1000 images per fold). In this case, the supervised training must strongly rely on the ability of the unsupervised method to learn discriminative features. Moreover, since the unlabeled dataset contains other types of animals (bears, rabbits, etc.) and vehicles (trains, buses, etc.) in addition to the ones in the labeled set, the unsupervised method should be able to generalize well.

To validate our method, we follow the experimental pipeline of \cite{Coates11_AISTATS}. We first extract random patches and normalize them for local brightness and contrast. Note that EPLS does not require any whitening of the input data, since it decorrelates the data during the training by means of the imposed strong sparsity properties of the output target. Then, we apply the system to retrieve sparse features of patches covering the input image, pool them into 4 quadrants and finally train a $L_2$ SVM for classification purposes. We tune the SVM parameter using 5-fold cross-validation. As in \cite{Ngiam11}, we use a receptive field of 10x10 pixels and a stride of 1. The number of outputs is set to $N_h = 1600$ for fair comparison with the other state-of-the-art methods. We also provide the results of our method with sign split ($N_h = 1600$x$2$, using $\mathbf{W}$ and $-\mathbf{W}$ for encoding as in \cite{Coates11_ICML}) and using the sparse coding (SC) encoder, which \cite{Coates11_ICML} found to be the best when small number of labeled data is available. For this encoder, we searched over the same set of parameter values as \cite{Coates11_ICML}, i.e., $\lambda = \{0.5, 0.75, 1.0, 1.25, 1.5\}$. The parameter $\lambda$ is tuned to consider the use of sparse coding as encoder after the training and, thus, does not belong to the method that we propose.

\begin{figure}
\centering
\includegraphics[width=1\linewidth]{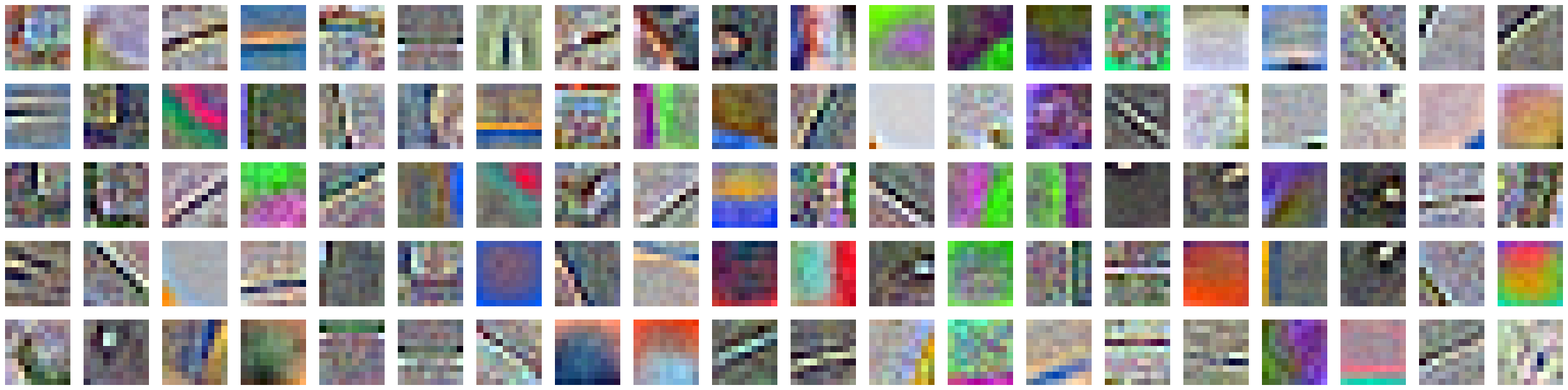}
\caption{Random subset of bases learned by EPLS, a receptive field of 10 pixels and $N_h = 1600$ (better seen in color).}
\label{fig:W}
\end{figure}

\begin{table}
\caption{Classification accuracy on STL-10.}
\footnotesize
\centering
\begin{tabular}{| c | c | c | c |}
\hline
\multicolumn{3}{|l|}{\textbf{Algorithm}} & \textbf{Accuracy}\\ \hline \hline
\multicolumn{4}{|l|}{\emph{Single-Layer \textbf{with} meta-parameters}} \\ \hline
\multicolumn{3}{|l|}{RICA \cite{Le11} ($1600$/Natural)} & 52.9\% \\ \hline
\multicolumn{3}{|l|}{OMP-1 ($1600$/Natural)} & 51.8\% (0.47\%)\\ \hline
\multicolumn{3}{|l|}{OMP-1 (whitening, $1600$/Natural)} & 53.1\% (0.52\%)\\ \hline
\multicolumn{3}{|l|}{OMP-1 (whitening, $1600$x$2$/Natural)} & 54.5\% (0.66\%)\\ \hline
\multicolumn{3}{|l|}{OMP-1 (whitening, $1600$x$2$/SC)} & 59.0\% (0.80\%) \\ \hline
\hline
\multicolumn{4}{|l|}{\emph{Single-Layer \textbf{without} meta-parameters}} \\ \hline
\multicolumn{3}{|l|}{Raw pixels} & 31.8\% (0.62\%) \\ \hline
\multicolumn{3}{|l|}{ICA (whitening, Complete/Natural)} & 48.0\% (1.47\%) \\ \hline
\multicolumn{3}{|l|}{K-means-tri (whitening, $1600$)} & 51.5\% (1.73\%)\\ \hline
\multicolumn{3}{|l|}{Sparse Filtering ($1600$/Natural)} & 53.5\% (0.53\%) \\ \hline \hline
& \begin{tabular}{@{}c@{}}Natural \\ ($1600$)\end{tabular} &  \begin{tabular}{@{}c@{}}Natural \\ ($1600$x$2$)\end{tabular} & \begin{tabular}{@{}c@{}}SC\\ ($1600$x$2$)\end{tabular} \\ \hline
EPLS & 56.6\% (0.66\%) & 56.9\% (0.50\%) & \textbf{61.0\% (0.58\%)}\\ \hline
\end{tabular}
\label{tab:stlresults}
\end{table}

Table \ref{tab:stlresults} summarizes the results obtained on this dataset compared to other state-of-the-art methods. When pairing each training method with its associated natural encoding, EPLS outperforms all the other methods. When pairing the training methods with sparse coding, EPLS outperforms the state-of-the-art best performer in single layer networks as well, achieving $61.0\%$ $(0.58\%)$ accuracy. Moreover, the standard deviation of the folds is lower than the one provided by OMP-1 with sparse coding encoding. Results are even more impressive if we compare them to meta-parameter free algorithms.

Figure \ref{fig:W} shows a subset of 100 randomly selected bases learned by our method, 10x10 pixel receptive field and a system of $N_h = 1600$ outputs. As shown in the figure, the method learns not only common bases such as oriented edges/ridges in many directions and colors but also corner detectors, tri-banded colored filters, center surrounds and Laplacian of Gaussians among others. This suggests that enforcing lifetime sparsity helps the system to learn a set of complex, rich and diversified bases.
\section{Computational complexity}
\label{subsec:cost}

The EPLS algorithm requires the computation of $\mathbf{T}$, which has $\mathcal{O}(NN_{h})$ cost, and therefore scales linearly on both $N$ and $N_{h}$. Since we can use vSGD for optimization, the method scales linearly on $N$ given a fixed number of epochs. Finally, applying the activation function, the cost of computing the derivative is linear with $N_{d}$, since we use a closed form for $\frac{\partial E}{\partial \Gamma}$.

The memory complexity is related to the mini-batch size $N_{b}$. Consequently, the method can scale gracefully to very large datasets: theoretically, it requires to store in memory the mini-batch input data $\mathbf{D}^{(b)}$ ($N_{b}N_d$ elements), output $\mathbf{H}^{(b)}$ ($N_{b}N_{h}$ elements), target $\mathbf{T}^{(b)}$ ($N_{b}N_{h}$ elements) and the system parameters to optimize $\Gamma$ ($N_h\left(N_d + 1\right)$ elements); a total amount of $N_h\left(N_d + 1\right) + N_b\left(N_d + 2N_h\right)$ elements.

\section{Discussion}
\label{sec:disc}

Our results show that simultaneously enforcing both population and lifetime sparsity helps in learning discriminative dictionaries, which reflect in better performance, especially when compared to meta-parameter free methods \cite{Ngiam11,Le11}. Experiments suggest that our algorithm is able to extract features that generalize well on unseen data. When comparing the performance STL-10 dataset, our algorithm outperforms state-of-the-art best performers. Results suggest that our algorithm helps the classifier in generalizing with a few training examples ($1\%$ of the dataset), gaining $2\%$ accuracy w.r.t. the state-of-the art best performer (OMP-1 paired with sparse coding) with a lower standard deviation across folds, suggesting more robustness to variations in the training folds.

It is important to highlight that OMP-1 can be seen as a special case of our algorithm, where the activation function is $| \mathbf{D}\mathbf{W} |$ and lifetime sparsity is not taken into account in the optimization process (potentially leading to dead outputs). Our algorithm has several advantages over OMP-1: (1) It can use any activation function; (2) by enforcing lifetime sparsity it does not suffer of the dead output problem, thus not requiring ad-hoc tricks to avoid it; (3) it does not require whitening, which can be a problem if the input dimensionality is large \cite{Le11}.

With our proposal, we advance in the meta-parameter free line of ICA \cite{Hyvarinen00} and sparse filtering \cite{Ngiam11}. It is clear that the advantage of sparse filtering over ICA comes from removing the orthogonality constraint, and imposing some sort of ``competition'' between outputs, which also permits overcomplete representations. Following this spirit, our algorithm imposes an even more strict form of competition to prevent dead outputs by means of Strong Lifetime Sparsity and confirms the trend of \cite{Ngiam11,Hyvarinen00} that data reconstruction seems not so important if the goal is to have a discriminative sparse system.

Last and most importantly, it is worth highlighting five interesting properties of the EPLS algorithm. First, the method is meta-parameter free, which highly simplifies the training process for practitioners, especially when used as a greedy pre-training method in deep architectures. Second, the method is fast and scales linearly with the number of training samples and the input/output dimensionalities. Third, EPLS is easy to implement. We implemented the EPLS in Algorithm \ref{algorithm:idealTarget} in less than 50 lines of C code. The mini-batch vSGD is a general purpose optimizer; our Matlab implementation of vSGD plus the EPLS mex source will be publicly available after publication. Fourth, the proposed learning strategy is not limited to perceptrons. Fifth, there is an interest in the literature in avoiding redundancy in the image representation by using the algorithms in a convolutional fashion \cite{Kavukcuoglu10}. For this purpose, the EPLS can be slightly modified to apply the procedure to a whole image at once and consider the mini-batch size to be the image divided into patches. This aspect is not considered in the paper and is left for future investigation.

\section{Conclusion}
\label{sec:concl}
In this paper, we introduced the Enforcing Population and Lifetime Sparsity method. The algorithm provides a \textbf{meta-parameter free, off-the-shelf, simple and computationally efficient} approach for unsupervised sparse feature learning. It seeks both lifetime and population sparsity in an explicit way in order to learn discriminative features, thus preventing dead outputs.

Results show that the method significantly outperforms all state-of-the-art methods on STL-10 dataset with lower standard deviation across folds, suggesting more robustness across training sets.

\bibliography{biblio}
\bibliographystyle{icml2014}

\end{document}